%% file: acl_latex.tex
\pdfoutput=1

\documentclass[11pt]{article}

\usepackage{booktabs}           
\usepackage{amsmath}            
\usepackage{graphicx,array}
\usepackage{multirow}
\usepackage{adjustbox}
\usepackage{tcolorbox}

\usepackage{acl}

\usepackage{times}
\usepackage{latexsym}

\usepackage[T1]{fontenc}

\usepackage[utf8]{inputenc}

\usepackage{microtype}

\usepackage{inconsolata}

\title{Convergences and Divergences between Automatic Assessment and Human Evaluation: Insights from Comparing ChatGPT-Generated Translation and Neural Machine Translation}

\author{
    Zhaokun Jiang$^\dagger$\quad Qianxi Lv$^\dagger$\quad Ziyin Zhang$^\ddagger$\thanks{\texttt{daenerystargaryen@sjtu.edu.cn}}\quad Lei Lei$^\diamond$ \\
    $^\dagger$ School of Foreign Languages, Shanghai Jiao Tong University\\
    $^\ddagger$ Department of Computer Science and Engineering, Shanghai Jiao Tong University\\
    $^\diamond$Institute of Corpus Studies and Applications, Shanghai International Studies University\\
}

\begin{document}
\maketitle
\begin{abstract}
Large language models have demonstrated parallel and even superior translation performance compared to neural machine translation (NMT) systems. However, existing comparative studies between them mainly rely on automated metrics, raising questions into the feasibility of these metrics and their alignment with human judgment. The present study investigates the convergences and divergences between automated metrics and human evaluation in assessing the quality of machine translation from ChatGPT and three NMT systems. To perform automatic assessment, four automated metrics are employed, while human evaluation incorporates the DQF-MQM error typology and six rubrics. Notably, automatic assessment and human evaluation converge in measuring formal fidelity (e.g., error rates), but diverge when evaluating semantic and pragmatic fidelity, with automated metrics failing to capture the improvement of ChatGPT's translation brought by prompt engineering. These results underscore the indispensable role of human judgment in evaluating the performance of advanced translation tools at the current stage.
\end{abstract}

\input{sec1_introduction-new}
\input{sec2-related}
\input{sec3-method}
\input{sec4-result}
\input{sec5-discussion}
\input{sec6-conclusion}

\bibliography{custom}

\onecolumn
\input{appendix}

\end{document}

%% file: sec1_introduction-new.tex
\section{Introduction}\label{sec:intro}

Translation quality assessment (TQA) for machine translation gains increasing prominence along with the development of AI-based translation technologies and the increasing popularity of machine translation. Researchers have dedicated extensive theoretical discussions to elucidating the evolving construct of translation quality~\citep{House2015}, and empirical investigations have led to the development of diverse assessment methods to keep abread with the latest machine translation tools~\citep{Eyckmans2016,Chung2020,2023Lu-Han,Bojar2016,Popovic2018,2018Bentivogli}. Yet, it leaves the exploration on the validity of automatic assessment less touched.

Along with the technological turn, neural machine translation (NMT) has witnessed remarkable advancements characterized by reduced grammatical errors and enhanced contextual comprehension~\citep{Gaspari2015,Cho2014}. NMT has been applied across non-literary and literary texts, showcasing satisfactory quality and even a certain degree of creativity~\citep{Hu-Li2023}. The advent of large language models, represented by ChatGPT~\citep{2020GPT3,2023GPT-4,2022PaLM,2023codesurvey}, has further accelerated the progress of translation automation by enabling interactive and customized experiences~\citep{Godwin-Jones2022}. When used for translation tasks, ChatGPT exhibits greater stylistic adaptability and contextual sensitivity in comparison to NMT systems~\citep{2023GPT-4}. This versatility is achieved through the use of written prompts, which are textual instructions to elicit desired outputs or responses from LLMs. Previous research has demonstrated that ChatGPT's translation performance can vary depending on the prompts employed~\citep{Hendy2023,Jiao2023,He2023}.

\begin{figure*}
    \centering
    \includegraphics[width=0.325\textwidth]{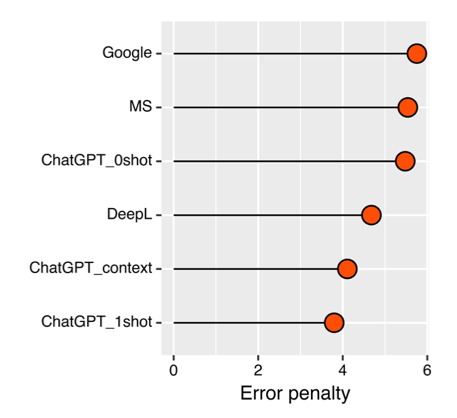}
    \includegraphics[width=0.33\textwidth]{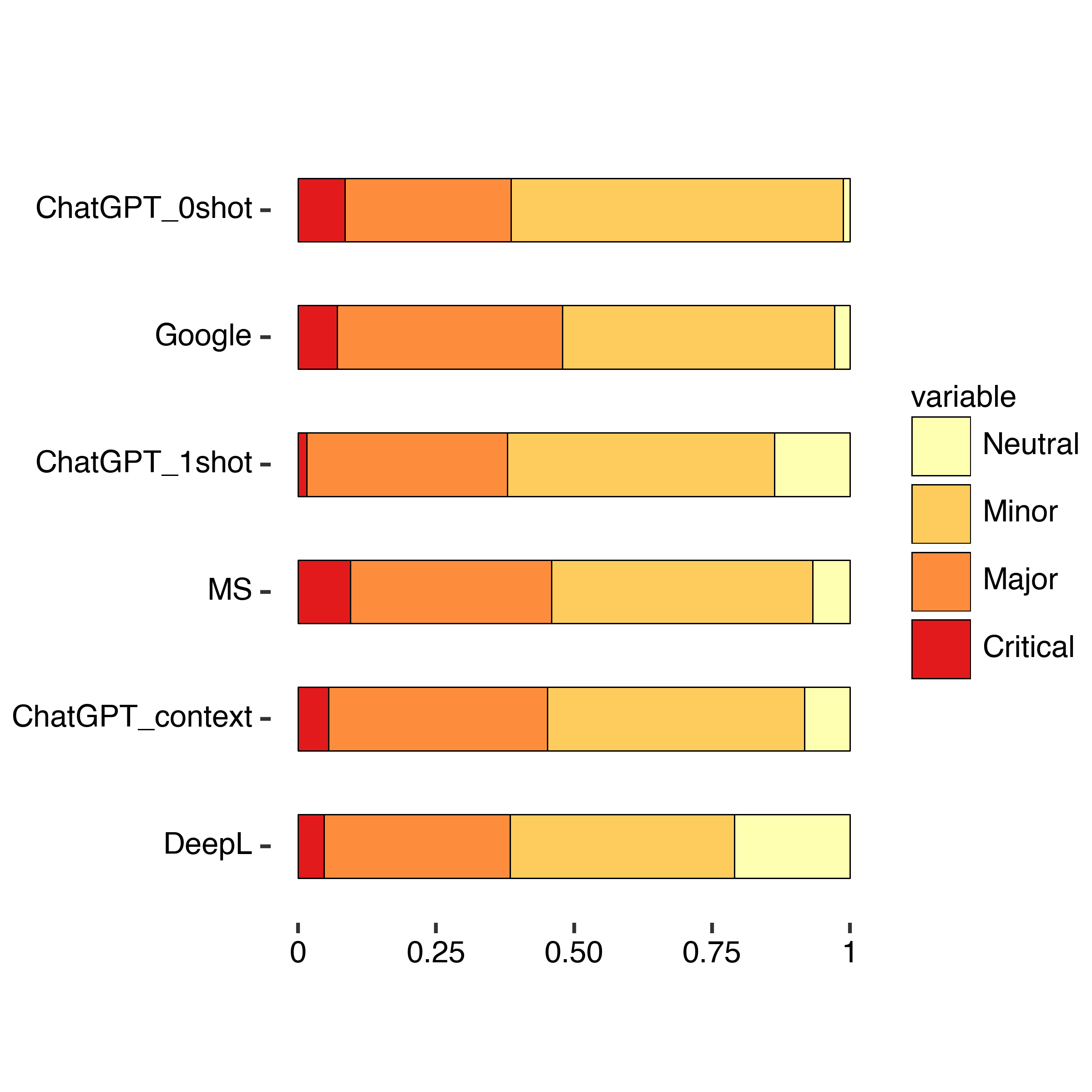}
    \includegraphics[width=0.33\textwidth]{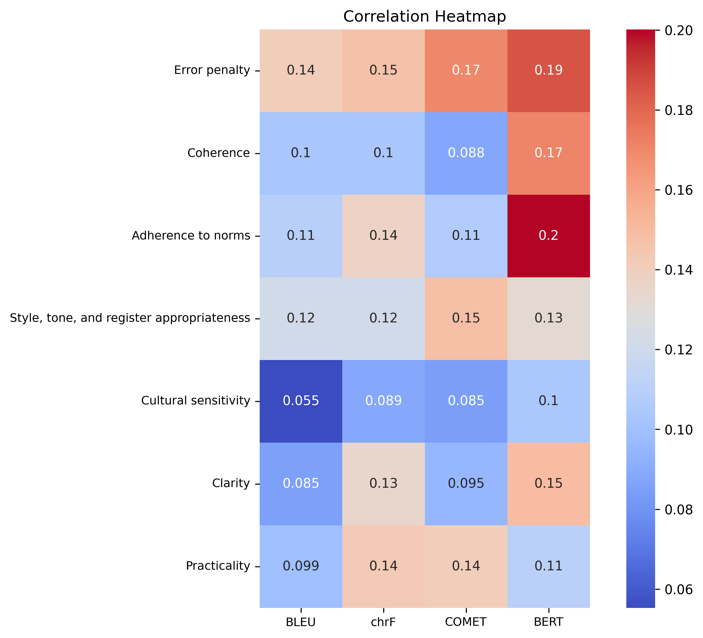}
    \caption{Total error penalty (left) and proportion of error severity (middle) assigned by human annotators to different translations. Right: correlation coefficients between human evaluation and automated metrics.}
    \label{fig:human-error-correlation}
\end{figure*}

In the context of large-scale and domain-specific MT tasks, automatic assessment has become the prevailing method for TQA due to its efficiency and standardization~\citep{Hendy2023}. To evaluate translation quality, a range of automated metrics have been developed. However, it is important to acknowledge that automatic assessment may over-emphasize quantifiable aspects of translation quality while downplaying other subtle yet vital dimensions associated with high-quality translation~\citep{OBrien2012,Pym-TorresSimón2021}. Specifically, most automated metrics mainly focus on accuracy, measuring the exact matching of n-grams (e.g., BLEU and chfF), or the extent to which the translation accurately reflects the semantic meaning of the reference (e.g., BERTScore and COMET). It is important to consider whether there are any essential dimensions that have been overlooked by automatic assessment and should be integrated into the existing automated methods.

As the counterpart of automatic assessment, human evaluation is widely regarded as the ``gold standard'' or the ``ground truth'', since human raters, especially professional translators, take factors such as cultural appropriateness, style, and norms into consideration. Human evaluation practices encompass a range of methods, including error analysis~\citep{Popovic2018} as well as emerging approaches like rubrics-referenced scoring~\citep{Han-Lu2023}, item-based assessment~\citep{Eyckmans2009}, and comparative judgment~\citep{Bojar2016}. However, in practice, human evaluation is rarely employed because it is time-consuming, cognitively-taxing, and labor-intensive~\citep{Han-Lu2023}, requiring scoring design, rater recruitment, rater training, and data collection. 

Hence, researchers have advocated for the integration of both approaches to overcome the limitations of human evaluation and leverage the strengths of automatic assessment. In pursuit of this objective, \citet{Callison-Burch2006} and \citet{Chung2020} conducted exploratory studies to examine the correlation between automated metrics and human scores. However, their investigations revealed relatively low and statistically insignificant correlation coefficients (below 0.3). These findings suggest that the automated scores are often unreliable and exhibit substantial variation across different contexts. Such inconsistencies may arise from the divergent constructs of translation quality. In other words, human raters may conceptualize translation quality differently from what can be quantified by automated metrics, and there may be significant dimensions within human evaluation that are not captured by automatic scores~\citep{Chung2020}. 

Understanding the disparity in how humans and automated systems perceive the construct of translation quality is the crucial first step to leverage their respective advantages. Ideally, given the esteemed status of human evaluation as the gold standard, it is essential to identify the distinctive dimensions inherent in human assessment and incorporate them into existing automated metrics.

The present study thus aims to address this issue by offering preliminary insights and potential solutions through empirical explorations. We conducted automatic assessment and human evaluation on translation outputs generated by three advanced NMT systems and ChatGPT. Special attention was paid to discern the dimensions that distinguish human-assigned scores from automatic scores Our investigation was conducted on a customized corpus consisting of 6,878 Spokesperson's Remarks. These diplomatic texts possess linguistic complexity and contextual intricacies, making them an ideal testing ground to evaluate how different machine translators handle the nuances and subtleties of language~\citep{2023Comparison}. By concentrating on this specific register, we aim to identify dimensions of translation quality that extend beyond accuracy, encompassing dimensions such as cultural awareness and textual coherence. Our ultimate objective is to make automatic assessment more human-like and fine-grained, and to develop a TQA method that exhibits enhanced validity, reliability, and applicability. This endeavor is particularly crucial in light of the emerging challenges posed by the continual advancements in large language models and NMT systems.

We constructed our major research questions (RQs) as follows:

\begin{enumerate}
    \item What do automatic assessment and human evaluation inform us about the translation quality of ChatGPT and NMT?
    \item To what degree do automated metrics align with human evaluation? 
    \item What are the unique insights from humans in evaluating the performance of advanced translation tools?
\end{enumerate}


%% file: sec2-related.tex
\section{Related Work}\label{sec:related}
\subsection{Improving the Translations of LLM via Prompt Engineering}
Large Language Models (LLMs) represented by OpenAI's ChatGPT have advanced remarkably in recent years, showcasing emergent abilities such as in-context learning and Chain-of-Thought reasoning~\citep{2020GPT3,2022PaLM,2022emergent,2022CoT,20220-shotCoT}, and several studies have explored the influence of prompting strategies on the translation performance of LLMs~\citep{Jiao2023,Hendy2023,Wang2023,Peng2023,Lu2023,Chen2023,He2023}. It has been found that with carefully written prompts, LLMs' translation quality can gain noticeable improvement, though such better performance is not seen in every scenario~\citep{Kocmi2023,Lu2023}. For instance, to inject cultural awareness into LLMs such as ChatGPT, \citet{Yao2023} propose ``cultural knowledge prompting''. Likewise, \citet{Mu2023} use translation memories as part of the prompts to enhance the translation performance of LLMs. 


\subsection{Translation Quality Assessment}
Much attention has been devoted to the field of Translation Quality Assessment (TQA) in the past few decades, with both empirical and theoretical discussions increasing markedly. Particularly, many studies have emphasized the importance of MT evaluation since MT has already become an integral instrument for translation practitioners, in both translator training and translation education. Coupled with this process is the rapid development of NMT engines and latelt the emergence of chatbots represented by ChatGPT, which poses great challenges to the applicability and validity of various assessment metrics and methods. 

Translation quality, however, has in fact no commonly agreed definition, as ``theorists and professionals overwhelmingly agree there is no single objective way to measure quality''~\citep{Drugan2013}. For those who share the belief that a translation should resemble the source text as closely as possible, quality is necessarily entangled with the concept of ``equivalence''. However, as translation studies experienced the cultural turn, this approach faces much criticism from the descriptive approach, which holds that emphasis should be laid on observing and describing the way translations are actually carried out, and how they are influenced by socio-cultural factors. The fit-for-purpose approach advocated by functionalists stresses that the notion of quality should be decided by how translations are received by end-users in real-life contexts. 

When MT becomes the subject of TQA, it can also draw on the above conceptions of quality. However, since MT is application-oriented in nature, calculability and measurability should be at the core of its quality assessment. One widely adopted view of MT quality is proposed by \citet{Koby2014}, which regards accuracy and fluency as the two determinants of translation quality. Alignment with human translation is also stressed, and \citet{Hassan2018} argue that only when MT shows no significant difference from human translations as measured by some scores can it be counted as high-quality translation. 

Despite the diversity of frameworks to address the evaluation of MT quality, we follow \citet{Chatzikoumi2020} to roughly divide approaches to TQA for MT into two lines: automated metrics and human evaluation. Automated metrics are of high efficiency, speed, consistency, and cost-effectiveness, and thus are widely adopted in both the academia and the industry. Within this line, metrics that compare a given translation to one or more reference translations are \textbf{reference-based}. They measure the similarity between the candidate MT and the reference(s) based on various linguistic and statistical features. Examples of reference-based metrics include BLEU (Bilingual Evaluation Understudy, \citealp{2002BLEU}), TER (Translation Edit Rate, \citealp{2006TER}), ChrF (Character n-gram F-score, \citealp{2015chrF}), and METEOR (Metric for Evaluation of Translation with Explicit Ordering, \citealp{2005METEOR}). 

\textbf{Reference-free} metrics, unlike reference-based metrics that rely on a set of high-quality reference translations, assess the quality of translations based solely on the characteristics of the translated text itself. These metrics aim to measure translation quality in a more independent and self-contained manner. They often focus on linguistic properties, statistical patterns, or other features of the translation, and are particularly useful in scenarios where high-quality reference translations are not readily available or when evaluating translations with specific requirements that do not align with existing references. Typical reference-free metrics include YiSi-2~\citep{2018YiSi}, COMET-QE~\citep{2021COMET-QE}, and QuestEval~\citep{Scialom2021}.

\textbf{Human evaluation} can also be categorized into different types based on the methodology used. In rubric scoring~\citep{2018Bentivogli}, human evaluators directly assign a score for translations within a fixed rating scale according to predefined criteria such as fluency, adequacy, accuracy, grammar, style, and overall coherence. Their judgments are typically combined to provide a final quality score. Another way of human evaluation is based on ranking~\citep{Bojar2016}, which involves comparing and ranking multiple translations to decide which is superior or more suitable for a given purpose or context. As for error-based evaluation, errors or issues present in translations are identified and categorized according to pre-determined error typology. Human evaluators analyze the translations and annotate specific errors, taking into consideration dimensions that include accuracy, fluency, terminology, style, and so on. The most representative error typologies are DQF (Dynamic Quality Framework) proposed by TAUS in 2011, MQM (Multidimensional Quality Metrics) developed by QTLaunchPad, and the harmonized DQF-MQM taxonomy~\citep{Popovic2018}. Finally, post-editing can also be used for evaluation purposes, since the posting-editing efforts by a human translator to render a MT ``good enough'' and ``deliverable'' implicitly reflects the quality of the raw translation~\citep{Massardo2016}. 

\subsection{Comparive Studies of LLM Translations and NMT}
As a promising contender to dedicated NMT systems, LLMs have attracted much attention from TQA-related studies that compare their performance against NMT engines using both automated metrics and human assessment. For example, \citet{Jiao2023} compare the translation performance of ChatGPT and GPT4 against three NMT systems: DeepL, Google Translate, and Tencent TranSmart using four automated metrics. They find that ChatGPT and GPT4 perform comparably well to NMT engines in specific European languages, but not in others. They also find that ChatGPT and GPT4 excel in translating spoken texts, but not in some specialized fields, such as abstracts of academic papers. Similarly, \citet{Hendy2023} show that for high-resource language pairs such as English and French, ChatGPT could exhibit state-of-the-art translation capabilities matching or even surpassing the mainstream NMT engines, while \citet{2023Comparison} extract a wide range of linguistic features of translations in political domains, and find that the translations of ChatGPT are closer to NMT than to human translations. \citet{Karpinska-Iyyer2023} show that when translating paragraph-level texts, ChatGPT produces fewer mistranslations, grammatical errors, and stylistic inconsistencies compared to Google Translate. 

The existing researches have confirmed the strong capacity of LLMs in translating high-resource languages such as English and German. However, we are not fully aware of their competence in understanding and translating middle and low-resource languages, such as Chinese. Also, most of the assessments are conducted on publicly available corpora such as WMT datasets, leaving more specialized translations - such as diplomatic discourse - as under-investigated domains awaiting more investigation~\citep{2023Comparison}.


%% file: sec3-method.tex
\section{Methodology}\label{sec:method}
\subsection{Corpus}
The corpus in this study consists of 6,878 pieces of Spokesperson's Remarks on regular press conferences, which contains questions proposed by foreign reporters and answers from the Chinese spokespersons centering several foreign affairs at a range of press conferences. The corpus includes 17,837 parallel sentences in total, which amount to 642K Chinese tokens and 450K English tokens. We focus on Chinese-to-English discourse out of three considerations: (1) data availability, since all the textual materials are easily found online; (2) high quality, as the human translation is performed by professional institutional translators; (3) relative complexity, because the sources texts contain many idiomatic expressions and political terms only seen in Chinese political contexts, which may pose challenge to NMT engines and ChatGPT. 

All of the questions are asked in English, and answers delivered by the spokespersons are in Chinese and then translated into English by institutional translators. Texts in our dataset are in fact transcripts of these press conferences, with some adjustments of wordings and contents, as well as corrections of speaking errors conducted by professional editors to be in line with the requirements of government websites. Since what the Chinese spokespersons express to the outside world concerns China's national interest and stance, their English translations are carefully curated and of premiere quality. All the materials can be directly downloaded from the official website of the Ministry of Foreign Affairs of the People's Republic of China\footnote{\url{https://www.mfa.gov.cn/eng/}}.

\begin{table*}[t]
\small
    \centering
    \begin{tabular}{lm{14cm}}
    \toprule
       \textbf{Method} & \textbf{Prompt} \\
    \midrule
        0-shot & Please translate the following Chinese sentence \{source\} into English. You should only output the translation.\vspace{0.05cm}\\\cline{2-2}
         & Please translate the following Chinese sentence \{source\} into English. You should only output the translation. \\
        \multirow{2}{*}{1-shot}& Here is an example for you: \\
        & \{source\} \\
        & \{reference\}\vspace{0.05cm}\\  \cline{2-2}
        Context & Please translate the following Chinese sentence \{source\} into English. It comes from a press conference in which a Chinese official interact with foreign reporters. The Chinese sentence is in the domain of politics. You should only output the translation.\\
    \bottomrule
    \end{tabular}
    \caption{Prompts for querying ChatGPT.}
    \label{tab:prompts}
\end{table*}

\subsection{Translation Tools}
For NMT systems, we consider Microsoft Translate, Google Translate, and DeepL. After the completion of machine translation, we conducted a manual examination of the outputs to verify that they were system errors. 

For LLM, we use GPT-3.5-Turbo, more commonly known as ChatGPT\footnote{We use gpt-3.5-turbo-0613 API, accessed on November 2nd, 2023.}. We handcraft three prompts: 0-shot, 1-shot, and a third prompt with additional information about the domain and context, as shown in Table~\ref{tab:prompts}.

\subsection{Evaluation Methods}
We adopt both automated metrics and human evaluation to balance efficiency and quality. Four widely-adopted automated metrics were selected. For human evaluation, we choose the integrated MQM-DFQ error typology and a six-dimensional analytic rubric scoring method.

\subsubsection{Automated Metrics}
\textbf{BLEU} involves a simple calculation of the number of n-gram matches between the MT output and one or multiple reference translations, with a penalty for unreasonably short translations. \textbf{ChrF} computes the F-score of character n-gram overlap instead. \textbf{BERTScore} leverages representations from pre-trained language models to compute the cosine similarity between translations and references, while \textbf{COMET} is similarly based on pre-trained language models, but also takes the source sentence into consideration during the scoring process.

We calculate BLEU, chrF, and BERTScore using torchmetrics\footnote{\url{https://github.com/Lightning-AI/torchmetrics}}. COMET is computed using Unbabel-COMET\footnote{\url{https://github.com/Unbabel/COMET}}. 

\subsubsection{Human Evaluation Based on Error Typology}
Multidimensional Quality Metrics (MQM) is a hierarchical and specifications-based framework for TQA~\citep{Lommel2013} that considers different dimensions of translation quality simultaneously in a comprehensive and systematic manner. Similarly, the DQF error typology developed by TAUS further contributes to this endeavor by providing a standardized classification system for identifying and categorizing errors in translations.

The similarity between MQM and DQF makes their marriage possible. The integrated DQF-MQM typology aims to provide a unified approach to TQA. It comprises seven primary error types, several subtypes under each primary type, and 4 severity levels (critical, major, minor, neutral) to address translation issues related to accuracy, locale convention, terminology, style, design, fluency, verity, and other issues. 

We choose accuracy, fluency, style, and terminology to form a subset of error typology to conduct human evaluation. To ensure the evaluation quality, annotators are provided with a detailed guideline, which can be found in the Appendix~\ref{sec:appendix-A}. It covers specifications of the textual materials, error typologies used in this study, severity levels, and penalty points. Our guideline is drafted in accordance with the official file provided by TAUS\footnote{\url{https://info.taus.net/dqf-mqf-error-typology-template-download}}. 

\subsubsection{Human Evaluation Based on Analytic Rubric Scoring}
Analytic rubric scoring is another method widely adopted in TQA research. It is founded on the assumption that the overall concept of quality can be broken down into individual components, and typically comprises several sub-scales addressing separate dimensions of translation~\citep{2023Lu-Han}. To complement the error typology-based evaluation, we propose six analytic rubrics to capture translation quality from different perspectives, encompassing dimensions of (1) coherence, (2) adherence to norms, (3) style, tone, and register appropriateness, (4) cultural sensitivity, (5) clarity, and (6) practicality. We use a 7-point Likert Scale and asked annotators to assign scores to the translation outputs for each rubric.

Considering time and costs, for each origin of translation (translations by ChatGPT under three prompts and by three NMT systems), we randomly sample 50 pieces of text to conduct human evaluation. To avoid pre-conceived judgements or biases on the quality of these translation tools, we do not inform the annotators of the origin of these translation samples beforehand. We recruited 6 annotators in total, all post-graduate students majoring in translation and interpreting (MTI). Each annotator is assigned 100 pieces of texts from all the six origins, and thus each text is annotated by two annotators. In terms of the error typology-based evaluation, the two annotations yield an average Cohen's Kappa of 0.73. Cases of disagreement (including the error type and the severity level) are settled through discussion and come to unanimous results. For the analytic rubric scoring, we average the scores assigned by the two annotators in each dimension.

\begin{table*}[t]
    \centering
    \begin{tabular}{l>{\raggedleft\arraybackslash}m{2cm}>{\raggedleft\arraybackslash}m{2cm}>{\raggedleft\arraybackslash}m{2cm}>{\raggedleft\arraybackslash}m{2cm}}
    \toprule
        Model & BLEU & Chrf & BERTScore & COMET\\
    \midrule
        ChatGPT 0-shot & 23.82 & 55.83 & 96.03 & 84.19\\
        ChatGPT 1-shot & 25.08 & 55.83 & 96.14 & 84.64\\
        ChatGPT with context & 23.98 & 56.19 & 96.38 & 85.30\\
        Google Translate & 25.28 & 55.86 & 96.12 & 83.02\\
        MS Translate & 29.16 & 59.17 & 96.15 & 84.41\\
        DeepL & 26.39 & 57.39 & 96.18 & 83.92 \\
    \bottomrule
    \end{tabular}
    \caption{Automated metrics of ChatGPT and NMT systems.}
    \label{tab:results-auto}
\end{table*}

\subsection{Analysis of Scores from Automated Metrics and Human Annotators}
To answer RQ1, we compute the average scores of the four automated metrics. Their descriptive statistics can be found in Appendix~\ref{sec:appendix-B}. To answer RQ2, we first provide annotation results guided by the MQM-DQF error typology, including the average error penalties, the frequencies of error types and subtypes, and the distribution of error severities. Then we show the results of analytic rubric scoring on six dimensions. The detailed descriptive statistics of human evaluation are listed in Appendix~\ref{sec:appendix-C}. To answer RQ3, we calculate the Pearson's correlation coefficients at the significance level of 0.05 to examine the inter-relationship between automated metrics and human scores. 

%% file: sec4-result.tex
\section{Results and Analysis}\label{sec:results}
This section provides results of automated metrics and human evaluation. Section~\ref{sec:result-auto} shows the performance of ChatGPT and NMT engines measured by four automated metrics. The first part of Section~\ref{sec:result-human} centers on the results of human evaluation, displaying error penalties, the distribution of error severities and error types in different translation outputs. The second part displays scores assigned by human annotators in six analytic rubrics. Section~\ref{sec:result-correlation} shows the inter-correlation between automated metrics and human evaluation.

\subsection{Automated Metrics}\label{sec:result-auto}
The results of ChatGPT and NMT systems' translations are presented in Table~\ref{tab:results-auto}. Compared with the 0-shot scenario, providing ChatGPT with an example or contextual information only brings slight increase in BLEU score. There is no noticeable change on other evaluative metrics, indicating that ChatGPT under the 0-shot condition has already demonstrated strong capability in handling this specific translation task. The high BERTScore and COMET scores prove that in terms of semantic accuracy, ChatGPT has no difficulty in understanding the characteristic expressions in the domain of diplomatic discourse, and can deliver relatively faithful English translations even under the 0-shot condition. Among the three NMT systems, MS Translate gains the highest scores on 3 out of 4 metrics, only slightly lagging behind DeepL on BERTScore. However, the differences among these systems are not marked.

\begin{table*}[t]
    \centering
    \begin{tabular}{llcccccc}
    \toprule
        &&\multicolumn{3}{c}{ChatGPT} & \multirow{2}{*}{Google} & \multirow{2}{*}{MS} & \multirow{2}{*}{DeepL} \\
        && 0-shot & 1-shot & w. context \\
    \midrule
        \multirow{6}{*}{Accuracy}& Addition & 1 & 0 & 1 & 0 & 0 & 0\\
        & Omission & 1 & 0 & 2 & 3 & 0 & 0\\
        & Mistranslation & 4 & 7 & 4 & 6 & 6 & 10\\
        & Over-translation & 1 & 1 & 1 & 2 & 2 & 2\\
        & Under-translation & 5 & 1 & 2 & 0 & 3 & 4\\
        \cline{2-8}
        & Total & 11 & 8 & 9 & 12 & 11 & 15\\
    \midrule
        \multirow{3}{*}{Fluency}& Punctuation & 0 & 0 & 1 & 2 & 0 & 0\\
        & Grammar & 6 & 5 & 8 & 5 & 8 & 8\\
        \cline{2-8}
        & Total & 6 & 5 & 8 & 7 & 8 & 8\\
    \midrule
        \multirow{3}{*}{Terminology}& Wrong terms & 9 & 6 & 6 & 7 & 4 & 7\\
        & Inconsistent usage & 0 & 0 & 0 & 0 & 0 & 0\\
        \cline{2-8}
        & Total & 9 & 7 & 7 & 7 & 5 & 8\\
    \midrule
        \multirow{4}{*}{Style}& Inconsistent style & 0 & 1 & 0 & 1 & 1 & 1\\
        & Awkward & 16 & 2 & 3 & 2 & 3 & 2\\
        & Unidiomatic & 3 & 13 & 7 & 9 & 4 & 9\\
        \cline{2-8}
        & Total & 19 & 17 & 12 & 13 & 9 & 13\\
    \midrule
        Other Errors &  & 0 & 1 & 0 & 1 & 0 & 0\\
    \bottomrule
    \end{tabular}
    \caption{Major types and subtypes of errors in different translations.}
    \label{tab:error-type}
\end{table*}
\begin{figure*}
    \centering
    \includegraphics[width=1\textwidth]{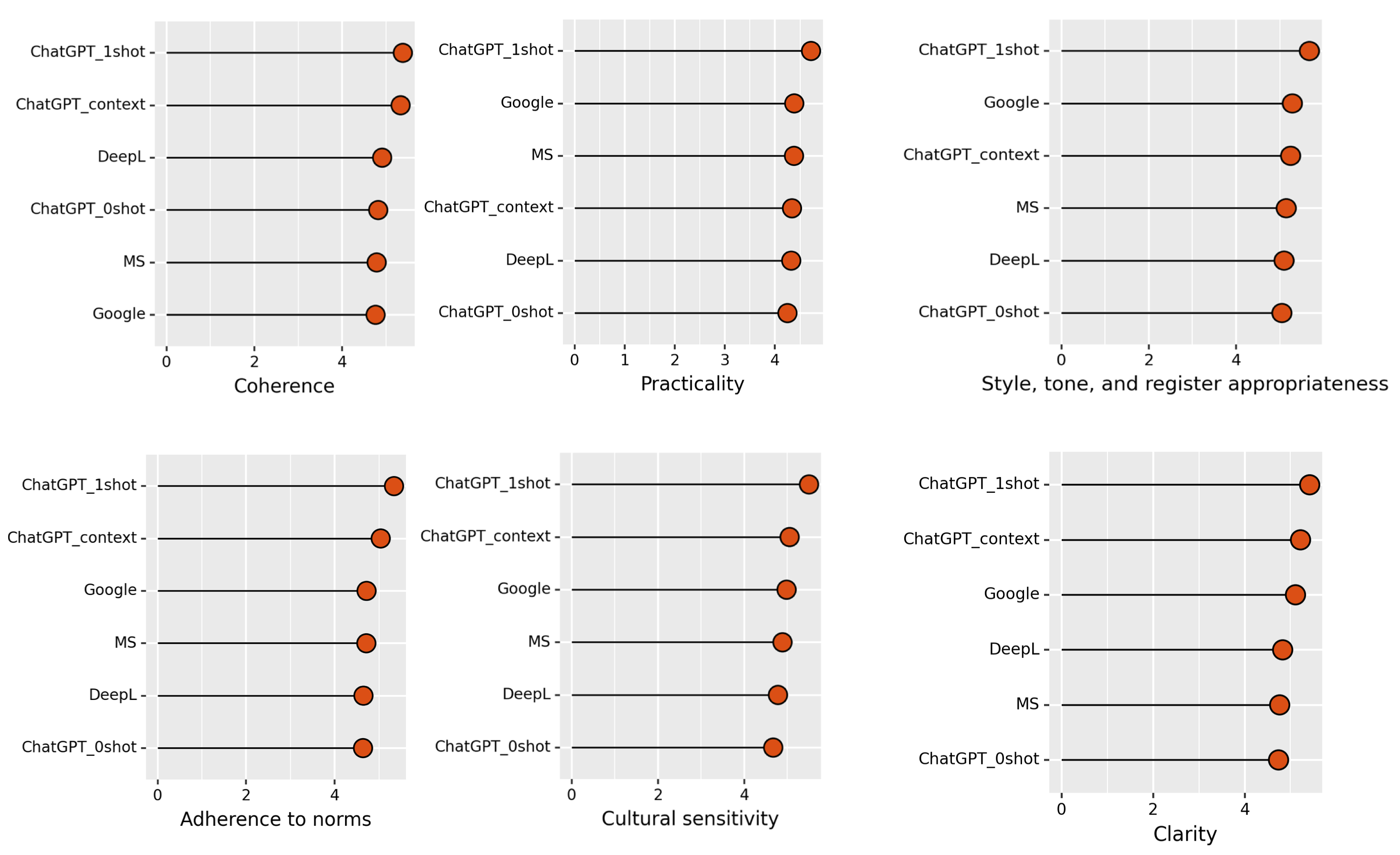}
    \caption{Human-assigned scores for each analytic rubric.}
    \label{fig:rubric}
\end{figure*}

\subsection{Human Assessment}\label{sec:result-human}
The total error penalty assigned by human annotators to each translation system is plotted in the left subfigure of Figure~\ref{fig:human-error-correlation}. We observe that ChatGPT under the 1-shot condition is assigned the lowest penalties, indicating its highest translation capacity among the translation outputs listed, while ChatGPT under the 0-shot condition performs much worse. This suggests that providing even a small amount of training data, in this case only one example, significantly improves the translation capability of ChatGPT. Error penalty in translations by ChatGPT when it is given contextual information is the second lowest, showing that incorporating context, possibly through the use of context-aware prompts, can improve the translation quality provided by ChatGPT. On the other hand, Google Translate, MS Translate, and DeepL have comparably high error penalties, indicating similar translation performance among these widely used machine translation systems. 

The middle subfigure of Figure~\ref{fig:human-error-correlation} displays the proportion of errors in each level of severity. Among NMT engines, DeepL contains the highest percentage of neutral errors and least percentage of major and critical errors, indicating that it demonstrates better capability compared with the other two NMT engines. In contrast, Google Translate sees the highest proportion of major errors in its translations. As for ChatGPT, most of its errors under the 0-shot condition are minor and major errors. Giving it an example significantly decreases the proportion of its critical errors and increases its neutral errors. Providing it with contextual information yields a similar effect.

Table~\ref{tab:error-type} shows the occurrence of errors under each major and sub-category. Overall, stylistic errors occur most frequently in most translations, followed by errors related to accuracy. While most errors of ChatGPT are attributed to style, accuracy poses the most challenge to NMT systems. Particularly, translations by ChatGPT are identified significantly more awkward and unidiomatic expressions compared with NMT systems.

In terms of analytic rubrics, Figure~\ref{fig:rubric} shows that ChatGPT under the 1-shot setting consistently receives the highest scores across all the rubrics, indicating its overall superiority under human evaluation. In contrast, ChatGPT in the 0-shot scenario ranks last in five out of six dimensions (adherence to norms, practicality, clarity, cultural sensitivity, as well as style, tone, and register appropriateness), which suggests that providing it with an example containing high-quality human translation contributes to boosting its translation performance markedly. Exposing ChatGPT to contextual information also serves to improve its translation quality in most cases (four out of six dimensions). The three NMT systems receive similar scores, showing that human annotators do not observe notable differences among their translation outputs.

\subsection{Correlation between Automated Metrics and Human Assessment}\label{sec:result-correlation}
Based on the results in Section~\ref{sec:result-auto} and ref~\ref{sec:result-human}, we calculate the correlation coefficients between human-assigned scores and scores from automated metrics. The results are shown in the right subfigure of Figure~\ref{fig:human-error-correlation}. We observe that in general, BERTScores align with human-assigned scores most closely, showing an average correlation coefficient of 0.16. BLEU scores deviate most from human evaluation, with a mere 0.10 correlation coefficient on average. However, the majority of these correlations are not statistically significant (the p values can be found in Appendix~\ref{sec:appendix-D}), suggesting that human evaluation and automated metrics tend to focus on different aspects of translation quality. In terms of aspects of human evaluation, error penalty shows the strongest correlation with automated scores in general, while cultural sensitivity demonstrates the weakest correlation.

%% file: sec5-discussion.tex
\section{Discussion}\label{sec:discussion}
From the results of automatic assessment, we can see that when measured by BLEU and Chrf, ChatGPT is overshadowed by NMT systems. This suggests that translations generated by ChatGPT tend to deviate more from the references in terms of the n-gram matches than NMT system, and ChatGPT may struggle to produce translations that precisely replicate the wording and phrasing of the references. However, its performance shines when evaluated using BERTScore and COMET score. These metrics take into account the semantic similarity and fluency of translations, rather than focusing solely on n-gram overlap. ChatGPT's ability to challenge or even surpass NMT systems when measured by these two metrics suggests that despite the deviations from the reference translations, its translations exhibit a strong semantic closeness to the references.

This observation aligns with previous researches on generative LLMs, which have shown that these models excel at capturing semantic relationships and generating coherent text. The underlying architecture of ChatGPT grants it exceptional context awareness and ability of language understanding, which enables its to produce translations that are more semantically aligned with the references, even if they may differ in specific word choices or phrase structure. It is worth noting that although BLEU and Chrf scores are widely adopted for evaluating translation quality, they have limitations, particularly when applied to language models like ChatGPT, which prioritize semantic coherence over exact phrasing. On the other hand, BERTScore and COMET provide a more nuanced evaluation by considering semantic aspects and fluency. These metrics are better at reflecting the strengths of ChatGPT in translating context-dependent texts. 

Another finding is that different prompting strategies do not exert significant boost to ChatGPT's translation performance, which can be attributed to two possible reasons. First, ChatGPT in the 0-shot scenario can already understand the semantic meaning of the original text perfectly, so that providing it with either extra information will not significantly affect its translation capability. Second, these automated metrics fail to capture the improvement brought by customized prompts. Drawing from results of human evaluation, the second explanation holds more feasibility, which will be further discussed later. 

In terms of human evaluation, we observe from the overall error penalties (see Figure~\ref{fig:human-error-correlation}) that ChatGPT tends to demonstrate strong plasticity in its translation performance: providing it with only one example or relevant contextual information can greatly reduce error penalties in its translation outputs and lead to a significant improvement in translation quality. 

This is further corroborated by evidence from scores of six analytic rubrics (see Figure~\ref{fig:rubric}), which demonstrates the variability of the performance of ChatGPT under different conditions. When exposed to a single example, ChatGPT consistently receives the highest scores across all the analytic rubrics, indicating its superior performance as evaluated by human annotators. In contrast, ChatGPT without any example (0-shot condition) ranks last in five out of six dimensions: adherence to norms, practicality, clarity, cultural sensitivity, and style, tone, and register appropriateness. This indicates that ChatGPT without the guidance of an example struggles in these areas and relies on the presence of a reference translation to enhance its performance. Similarly, exposing ChatGPT to contextual information improves its translation quality across all the six rubrics, showing that the inclusion of context aids ChatGPT in generating more contextually appropriate translations. This informs us the importance of crafting appropriate and relevant prompts to fully uncover the potential of ChatGPT in generating high-quality translations. On the other hand, the three NMT systems receive similar scores across the assessed dimensions, implying that human annotators do not observe significant differences in translation quality among these NMT systems.

In addition, NMT engines are more prone to mistakes compared with ChatGPT. This corresponds to previous findings reported by \citet{Manakhimova2023}, who investigate the ability of GPT4 and NMT systems in handling challenging translation issues. Their study shows that GPT4 outperforms NMT engines in most cases, displaying fewer mistakes and higher accuracy. This is further supported by the distribution of errors (see Table~\ref{tab:error-type}), which shows that accuracy poses the greatest difficulty for NMT systems in general. 

In contrast, ChatGPT exhibits style errors as its primary challenge. In particular, translations by ChatGPT are noted for containing more awkward and idiomatic expressions compared to NMT systems, but providing relevant context notably reduced its style-related errors. ChatGPT's style errors may stem from its language generation capabilities, as it tends to prioritize semantic coherence over precise phrasing, leading to stylistic variations. The awkward and idiomatic expressions may arise from its tendency to generate creative and diverse outputs, which can sometimes result in less conventional or less fluent translations.

Finally, the Pearson's correlations between human-assigned scores and automated metrics are weak and non-significant, though some specific dimensions show slightly higher correlations than others. This is somehow contradictory to findings in \citet{2023Lu-Han}, possibly because assessing interpreting is different from assessing translation, with the latter having higher requirements on aspects other than accuracy. Our results highlight the challenges in relying solely on automated scores to assess translation quality, as they may not fully capture the intricacies of dimensions such as adherence to norms, cultural sensitivity, clarity, and practicality. Human evaluation as a more contextual and culture-specific way of assessment is necessary to obtain a comprehensive understanding of translation quality. 

What implications do these results have for future research in TQA? First, we should acknowledge the limitations of traditional metrics. Our findings highlight that metrics like BLEU and chrF, which primarily focus on n-gram overlap and exact phrasing, cannot effectively capture the superior performance of language models such as ChatGPT in a range of critical dimensions for high-quality human-like translations. This calls for the development and adoption of more nuanced evaluation metrics that consider cultural aspects and contextual appropriateness. Next, carefully crafted and customized prompts are needed to fully unleash the great potential of ChatGPT as a capable machine translator. Our results show that providing a single example or some relevant contextual information can greatly reduce its errors and drive up its scores in all the analytic rubrics. This highlights the practical significance of tailoring prompts to guide the generation process and enhance the translation quality of ChatGPT. 

%% file: sec6-conclusion.tex
\section{Conclusion}\label{sec:conclusion}
This study compares the translation quality of ChatGPT and three NMT engines by using both automated metrics and human evaluation. For the former, we compute four widely-adopted metrics - BLEU, chrF, COMET, and BERTScore - and find that they fail to distinguish high-quality and lower-quality translations. For the latter, we conduct annotation based on both the integrated MQM-DQF error typology and six analytic rubrics. Results show that exposing ChatGPT to one example or context-relevant information greatly boosts its performance under all dimensions of human evaluation. To examine the correlation between automated metrics and human evaluation, we calculate the pairwise Pearson’s correlation coefficients. The weak and non-significant results overall demonstrate that human understanding of translation quality is significantly different from what is captured by automated metrics.  

In light of the findings discussed above, we suggest two directions to advance researches in TQA: Firstly, further exploration is needed to develop more effective evaluation metrics that can better capture the nuances of translation quality, particularly in terms of coherence, clarity, practicality adherence to norms, cultural sensitivity, as well as appropriateness of style, tone, and register. Secondly, it is crucial to continue investigating the role of proper prompts and contextual information in improving the performance of language models like ChatGPT, which exhibit strong context awareness and language understanding capabilities.

%% file: appendix.tex
\appendix

\section{Guidelines for Human Assessment under the MQM Error Typology}\label{sec:appendix-A}
\subsection{General Instructions}
\begin{tcolorbox}
You will be assessing translations at the sentence level. Each translated sentence is aligned with its corresponding source text. You have the flexibility to revise previous annotations as needed.

\textbf{There are two tasks for you to finish.}

\textbf{The first one is error-analysis-based.} Your task is to identify errors within each translated sentence, with a maximum limit of five errors. If there are more than five errors, focus on marking the five most severe ones. In cases where the translation is severely distorted or unrelated to the source, mark a single Non-translation error that covers the entire segment.

To identify an error, highlight the relevant portion of the translation and choose a category/sub-category and severity level from the available options. When identifying errors, please be as fine-grained as possible. For instance, if a sentence contains two mistranslated words, record them as separate mistranslation errors. If a single section of text has multiple errors, indicate the most severe one.

Please pay particular attention to the context when annotating. If a translation may be questionable on its own but fits within the context, it should not be considered erroneous. Conversely, if a translation might be acceptable in some contexts, but not for the current sentence, mark it as incorrect.

There is a special error category called Non-translation, which can only be used once per sentence and should encompass the entire sentence. If Non-translation is selected, no other errors should be identified. 

\textbf{The second task is impression-based.} You need to report your level of agreement with 7 statements on a 7-point Likert scale, in which 1 means “completely disagree”, and 7 means “completely agree”. The statements are the following:
\begin{enumerate}
    \item \textbf{Coherence}: The translation flow is mostly smooth and coherent. There is no logical disconnection or meaning inconsistency.
    \item \textbf{Adherence to norms}: The translation fulfills the common standards, requirements, and norms of political translation.
    \item \textbf{Style, tone, and register appropriateness}: The translation is consistent in style, tone, and register with the source text.  For example, if the source text has a formal tone and sophisticated style, the translation also reflects that formality and sophistication.
    \item \textbf{Cultural sensitivity}: The translation demonstrates cultural sensitivity. It suitably conveys culture specific items (CSI), humor, and other cultural nuances in a way that is understandable and relatable to the target audience. 
    \item \textbf{Clarity}: The translation is clear and easily understandable to the target audience. It does not contain ambiguities, excessive jargon, or overly complex language that may hinder comprehension.
    \item \textbf{Practicality}: The translation can be directly put for actual use. In this case, it can be put on the government website for people across the world to read.
\end{enumerate}
\end{tcolorbox}

\subsection{Specifications}
We select a portion of parameters from the ASTM F2575-14 speciﬁcations\footnote{\url{https://www.astm.org/f2575-14.html}} to describe what is expected of the translation.

\subsubsection*{Source-content information}

\begin{tcolorbox}
Source language: Chinese

Text type: political and diplomatic texts remarks from the Chinese spokesmen.

Audience: Chinese readers who intend to know China's stance on a range of important foreign affairs (e.g., reporters, politicians, and diplomats). 

Purpose: to deliver China's stance and attitudes on a range of important foreign affairs.

Volume: you will be given 100 sentences.

Complexity: usually written in a relatively complex and formal style, commonly found in official statements or diplomatic contexts. 

Origin: official website of the Ministry of Foreign Affairs of The People's Republic of China (https://www.fmprc.gov.cn/fyrbt\_673021/dhdw\_673027/index\_1.shtml) 
\end{tcolorbox}

\subsubsection*{Target content requirements}

\begin{tcolorbox}
Target language: English

Audience: international readers who intend to know China's stance on a range of important foreign affairs (e.g., reporters, politicians, and diplomats).

Purpose: to deliver China's stance and attitudes on a range of important foreign affairs.

Format: written texts displayed on the government website, which are transcribed and carefully edited from spoken remarks.

Style: in a formal, official, and often assertive tone commonly found in official statements or diplomatic discourse.
\end{tcolorbox}

\subsection{Error Typology, Severity Levels and Penalty Levels}
\subsubsection*{Error Typology}
\begin{tcolorbox}
\adjustbox{width=\textwidth,center}{
\begin{tabular}{llm{10cm}}
\toprule
    Error Type & Subtype & Definition \\
\midrule
    \multirow{7}{*}{Accuracy}  & & The target text does not accurately reflect the source text, allowing for any differences authorized by specifications.\\\cline{2-3}
    & Addition & The target text includes text not present in the source. \\
    & Omission & Content is missing from the translation that is present in the source. \\
    & Mistranslation & The target content does not accurately represent the source content. \\
    & Over-translation & The target text is more specific than the source text. \\
    & Under-translation & The target text is less specific than the source text.\\
\midrule
    \multirow{8}{*}{Fluency} && Issues related to the form or content of a text, irrespective as to whether it is a translation or not.\\\cline{2-3}
    & Punctuation & is used incorrectly (for the locale or style).\\
    & Spelling & Issues related to spelling of words.\\
    & Grammar & Issues related to the grammar or syntax of the text, other than spelling and orthography.\\
    & Inconsistency & The text shows internal inconsistency.\\
    & Link/cross-reference & Links are inconsistent in the text.\\
\midrule
    \multirow{4}{*}{Terminology} & & A term (domain-specific word) is translated with a term other than the one expected for the domain or otherwise specified.\\\cline{2-3}
    & Wrong terms & Use of term that it is not the term a domain expert would use or because it gives rise to a conceptual mismatch.\\
    & Inconsistent use of terminology & Terminology is used in an inconsistent manner within the text.\\
\midrule
    \multirow{4}{*}{Style} && The text has stylistic problems.\\\cline{2-3}
    & Inconsistent style & Style is inconsistent within a text.\\
    & Awkward & A text is written with an awkward style.\\
    & Unidiomatic & The content is grammatical, but not idiomatic.\\
\midrule
    Other && Any other issues.\\
\bottomrule
\end{tabular}
}
\end{tcolorbox}

\subsubsection*{Severity Levels}
\begin{tcolorbox}
\adjustbox{width=\textwidth,center}{
\begin{tabular}{lm{12cm}}
\toprule
Non-translation & The translation is severely distorted or unrelated to the source.\\\cline{2-2}
Critical & Errors that may carry health, safety, legal or financial implications, violate geopolitical usage guidelines, damage the entities' reputation, or which could be seen as offensive. \\\cline{2-2}
Major & Errors that may confuse or mislead the readers due to significant change in meaning or because errors appear in a visible or important part of the content.\\\cline{2-2}
Minor & Errors that don't lead to loss of meaning and wouldn't confuse or mislead the readers but would be noticed, would decrease stylistic quality, fluency or clarity, or would make the content less appealing.\\\cline{2-2}
Neutral & Used to log additional information, problems or changes to be made that don't count as errors, e.g., they reflect a reviewer's choice or preferred style, they are repeated errors or instruction/glossary changes not yet implemented, a change to be made that the translator is not aware of.\\
\bottomrule
\end{tabular}
}
\end{tcolorbox}

\subsubsection*{Penalty Levels}
\begin{tcolorbox}
\centering
\begin{tabular}{lll}
\toprule
Non-translation & 100 & Deduct the penalty points for non-translation errors\\
Critical errors & 25 & Deduct the penalty points for critical errors\\
Major errors & 10 & Deduct the penalty points for major errors\\
Minor errors & 1 & Deduct the penalty points for minor errors\\
Neutral errors & 0 & Deduct the penalty points for neutral errors\\
\bottomrule
\end{tabular}
\end{tcolorbox}

\newpage
\section{Descriptive Statistics of Scores Using Automated Metrics}\label{sec:appendix-B}
The descriptive statistics of automated metrics (BLEU, chrF, COMET, and BERTScore) are given in Table~\ref{tab:stat-bleu}, \ref{tab:stat-chrf}, \ref{tab:stat-comet}, and \ref{tab:stat-bertscore} respectively.
\begin{table}[h]
    \centering
    \begin{tabular}{lrrrrrr}
    \toprule
         & Mean & Std. & Min & Max & Kurtosis & Skewness\\
    \midrule
        ChatGPT 0-shot & 0.228 & 0.196 & 0.000 & 1.000 & 0.211 & 0.605\\
        ChatGPT 1-shot & 0.251 & 0.213 & 0.000 & 1.000 & 0.317 & 0.662\\
        ChatGPT w. context & 0.235 & 0.200 & 0.000 & 1.000 & 0.172 & 0.607 \\
        Google Translate & 0.243 & 0.209 & 0.000 & 1.000 & 0.306 & 0.654\\
        MS Translate & 0.264 & 0.217 & 0.000 & 1.000 & -0.089 & 0.528\\
        DeepL & 0.292 & 0.236 & 0.000 & 1.000 & -0.039 & 0.562 \\
    \bottomrule
    \end{tabular}
    \caption{BLEU statistics.}
    \label{tab:stat-bleu}
\end{table}
\begin{table}[h]
    \centering
    \begin{tabular}{lrrrrrr}
    \toprule
         & Mean & Std. & Min & Max & Kurtosis & Skewness\\
    \midrule
        ChatGPT 0-shot & 0.558 & 0.152 & 0.076 & 1.000 & 0.214 & -0.333\\
        ChatGPT 1-shot & 0.586 & 0.092 & 0.173 & 1.000 & 0.357 & -0.178\\
        ChatGPT w. context & 0.563 & 0.154 & 0.067 & 1.000 & 0.215 & -0.314\\
        Google Translate & 0.559 & 0.167 & 0.000 & 1.000 & 0.754 & -0.421 \\
        MS Translate & 0.592 & 0.169 & 0.065 & 1.000 & 0.039 & -0.147 \\
        DeepL & 0.574 & 0.163 & 0.065 & 1.000 & 0.077 & -0.282\\
    \bottomrule
    \end{tabular}
    \caption{ChrF statistics.}
    \label{tab:stat-chrf}
\end{table}
\begin{table}[h]
    \centering
    \begin{tabular}{lrrrrrr}
    \toprule
         & Mean & Std. & Min & Max & Kurtosis & Skewness\\
    \midrule
        ChatGPT 0-shot & 0.842 & 0.059 & 0.441 & 0.986 & 5.068 & -1.512\\
        ChatGPT 1-shot & 0.846 & 0.061 & 0.441 & 0.986 & 3.902 & -1.296\\
        ChatGPT w. context & 0.841 & 0.062 & 0.368 & 0.986 & 6.591 & -1.714\\
        Google Translate & 0.830 & 0.084 & 0.269 & 0.986 & 13.968 & -2.824\\
        MS Translate & 0.844 & 0.068 & 0.341 & 0.986 & 4.771 & -1.386\\
        DeepL & 0.839 & 0.064 & 0.441 & 0.986 & 3.549 & -1.166\\
    \bottomrule
    \end{tabular}
    \caption{COMET statistics.}
    \label{tab:stat-comet}
\end{table}
\begin{table}[!h]
    \centering
    \begin{tabular}{lrrrrrr}
    \toprule
         & Mean & Std. & Min & Max & Kurtosis & Skewness\\
    \midrule
        ChatGPT 0-shot & 0.960 & 0.007 & 0.938 & 0.992 & 0.394 & 0.302\\
        ChatGPT 1-shot & 0.961 & 0.008 & 0.940 & 0.999 & 0.992 & 0.579\\
        ChatGPT w. context & 0.961 & 0.008 & 0.938 & 0.995 & 0.470 & 0.471 \\
        Google Translate & 0.961 & 0.008 & 0.939 & 0.995 & 0.313 & 0.328\\
        MS Translate & 0.962 & 0.008 & 0.938 & 0.996 & 0.832 & 0.540\\
        Tengxun Translate & 0.962 & 0.008 & 0.939 & 0.995 & 0.283 & 0.408\\
    \bottomrule
    \end{tabular}
    \caption{BERTScore statistics.}
    \label{tab:stat-bertscore}
\end{table}

\newpage
\section{Descriptive Statistics of Scores Assigned by Human Annotators}\label{sec:appendix-C}
The descriptive statistics of annotators' assigned scores to the six systems are given in Table~\ref{tab:stat-human-gpt0}-\ref{tab:stat-human-deepl}.

\begin{table}[h]
    \centering
    \adjustbox{width=\textwidth+1cm,center}{
    \begin{tabular}{l>{\raggedleft\arraybackslash}m{1.8cm}>{\raggedleft\arraybackslash}m{1.8cm}>{\raggedleft\arraybackslash}m{1.8cm}>{\raggedleft\arraybackslash}m{1.8cm}>{\raggedleft\arraybackslash}m{1.8cm}>{\raggedleft\arraybackslash}m{1.8cm}>{\raggedleft\arraybackslash}m{1.8cm}}
    \toprule
         & Error penalty & Coherence & Adherence to norms & Style, tone, and register appropriateness & Cultural sensitivity & Clarity & Practicality\\
    \midrule
        Mean & -4.110 & 4.830 & 4.890 & 5.250 & 4.850 & 4.970 & 4.340\\
        Std. & -6.302 & 1.640 & 1.406 & 1.351 & 1.591 & 1.611 & 1.713\\
        Max & 10.000 & 2.000 & 2.000 & 2.000 & 1.000 & 1.000 & 1.000 \\
        Min & -35.000 & 7.000 & 7.000 & 7.000 & 7.000 & 7.000 & 7.000 \\
        Kurtosis & -4.373 & -0.973 & -1.028 & -0.603 & -0.917 & -0.728 & -1.066\\
        Skewness & -1.225 & -0.444 & -0.023 & -0.560 & -0.357 & -0.563 & -0.101 \\
    \bottomrule
    \end{tabular}
    }
    \caption{Human annotation statistics of ChatGPT 0-shot.}
    \label{tab:stat-human-gpt0}
\end{table}
\begin{table}[h]
    \centering
    \adjustbox{width=\textwidth+1cm,center}{
    \begin{tabular}{l>{\raggedleft\arraybackslash}m{1.8cm}>{\raggedleft\arraybackslash}m{1.8cm}>{\raggedleft\arraybackslash}m{1.8cm}>{\raggedleft\arraybackslash}m{1.8cm}>{\raggedleft\arraybackslash}m{1.8cm}>{\raggedleft\arraybackslash}m{1.8cm}>{\raggedleft\arraybackslash}m{1.8cm}}
    \toprule
         & Error penalty & Coherence & Adherence to norms & Style, tone, and register appropriateness & Cultural sensitivity & Clarity & Practicality\\
    \midrule
        Mean & -3.808 & 5.380 & 5.340 & 5.680 & 5.470 & 5.420 & 4.717 \\
        Std. & -7.675 & 1.285 & 1.121 & 1.024 & 1.259 & 1.365 & 1.385\\
        Max & 10.000 & 1.000 & 2.000 & 2.000 & 2.000 & 2.000 & 1.000 \\
        Min & -25.000 & 7.000 & 7.000 & 7.000 & 7.000 & 7.000 & 7.000 \\
        Kurtosis & -1.268 & 0.708 & -0.138 & 0.297 & -0.233 & -0.413 & -0.303\\
        Skewness & -1.218 & -0.937 & -0.438 & -0.578 & -0.604 & -0.598 & -0.386\\
    \bottomrule
    \end{tabular}
    }
    \caption{Human annotation statistics of ChatGPT 1-shot.}
    \label{tab:stat-human-gpt1}
\end{table}
\begin{table}[h]
    \centering
    \adjustbox{width=\textwidth+1cm,center}{
    \begin{tabular}{l>{\raggedleft\arraybackslash}m{1.8cm}>{\raggedleft\arraybackslash}m{1.8cm}>{\raggedleft\arraybackslash}m{1.8cm}>{\raggedleft\arraybackslash}m{1.8cm}>{\raggedleft\arraybackslash}m{1.8cm}>{\raggedleft\arraybackslash}m{1.8cm}>{\raggedleft\arraybackslash}m{1.8cm}}
    \toprule
         & Error penalty & Coherence & Adherence to norms & Style, tone, and register appropriateness & Cultural sensitivity & Clarity & Practicality\\
    \midrule
        Mean & -4.110 & 4.830 & 4.890 & 5.250 & 4.850 & 4.970 & 4.340\\
        Sd. & -6.302 & 1.640 & 1.406 & 1.351 & 1.591 & 1.611 & 1.713\\
        Max & 10.000 & 2.000 & 2.000 & 2.000 & 1.000 & 1.000 & 1.000 \\
        Min & -35.000 & 7.000 & 7.000 & 7.000 & 7.000 & 7.000 & 7.000 \\
        Kurtosis & -4.373 & -0.973 & -1.028 & -0.603 & -0.917 & -0.728 & -1.066\\
        Skewness & -1.225 & -0.444 & -0.023 & -0.560 & -0.357 & -0.563 & -0.101 \\
    \bottomrule
    \end{tabular}
    }
    \caption{Human annotation statistics of ChatGPT with context.}
    \label{tab:stat-human-gptc}
\end{table}
\begin{table}[h]
    \centering
    \adjustbox{width=\textwidth+1cm,center}{
    \begin{tabular}{l>{\raggedleft\arraybackslash}m{1.8cm}>{\raggedleft\arraybackslash}m{1.8cm}>{\raggedleft\arraybackslash}m{1.8cm}>{\raggedleft\arraybackslash}m{1.8cm}>{\raggedleft\arraybackslash}m{1.8cm}>{\raggedleft\arraybackslash}m{1.8cm}>{\raggedleft\arraybackslash}m{1.8cm}}
    \toprule
         & Error penalty & Coherence & Adherence to norms & Style, tone, and register appropriateness & Cultural sensitivity & Clarity & Practicality\\
    \midrule
        Mean & -5.660 & 5.260 & 5.220 & 5.290 & 5.170 & 5.110 & 4.384\\
        Std. & -7.907 & 1.474 & 1.345 & 1.328 & 1.464 & 1.645 & 1.658\\
        Max & 5.000 & 1.000 & 2.000 & 3.000 & 2.000 & 1.000 & 1.000 \\
        Min & -25.000 & 7.000 & 7.000 & 7.000 & 7.000 & 7.000 & 7.000 \\
        Kurtosis & -0.210 & 0.215 & -0.980 & -1.116 & -1.221 & -0.754 & -0.890\\
        Skewness & -1.066 & -0.911 & -0.231 & -0.179 & -0.277 & -0.588 & -0.058\\
    \bottomrule
    \end{tabular}
    }
    \caption{Human annotation statistics of Google Translate.}
    \label{tab:stat-human-google}
\end{table}
\begin{table}[h]
    \centering
    \adjustbox{width=\textwidth+1cm,center}{
    \begin{tabular}{l>{\raggedleft\arraybackslash}m{1.8cm}>{\raggedleft\arraybackslash}m{1.8cm}>{\raggedleft\arraybackslash}m{1.8cm}>{\raggedleft\arraybackslash}m{1.8cm}>{\raggedleft\arraybackslash}m{1.8cm}>{\raggedleft\arraybackslash}m{1.8cm}>{\raggedleft\arraybackslash}m{1.8cm}}
    \toprule
         & Error penalty & Coherence & Adherence to norms & Style, tone, and register appropriateness & Cultural sensitivity & Clarity & Practicality\\
    \midrule
        Mean & -5.540 & 4.788 & 4.717 & 5.152 & 4.889 & 4.768 & 4.384\\
        Std. & -9.440 & 1.728 & 1.597 & 1.466 & 1.684 & 1.778 & 1.748\\
        Max & 10.000 & 1.000 & 1.000 & 1.000 & 1.000 & 1.000 & 1.000 \\
        Min & -35.000 & 7.000 & 7.000 & 7.000 & 7.000 & 7.000 & 7.000 \\
        Kurtosis & -1.262 & -0.959 & -0.502 & 0.210 & -0.554 & -0.868 & -1.055\\
        Skewness & -1.407 & -0.397 & -0.452 & -0.752 & -0.559 & -0.489 & -0.358\\
    \bottomrule
    \end{tabular}
    }
    \caption{Human annotation statistics of Microsoft Translate.}
    \label{tab:stat-human-ms}
\end{table}
\begin{table}[!h]
    \centering
    \adjustbox{width=\textwidth+1cm,center}{
    \begin{tabular}{l>{\raggedleft\arraybackslash}m{1.8cm}>{\raggedleft\arraybackslash}m{1.8cm}>{\raggedleft\arraybackslash}m{1.8cm}>{\raggedleft\arraybackslash}m{1.8cm}>{\raggedleft\arraybackslash}m{1.8cm}>{\raggedleft\arraybackslash}m{1.8cm}>{\raggedleft\arraybackslash}m{1.8cm}}
    \toprule
         & Error penalty & Coherence & Adherence to norms & Style, tone, and register appropriateness & Cultural sensitivity & Clarity & Practicality\\
    \midrule
        Mean & -4.900 & 4.950 & 4.677 & 5.121 & 4.790 & 4.830 & 4.300\\
        Std. & -6.870 & 1.438 & 1.268 & 1.118 & 1.266 & 1.443 & 1.501 \\
        Max & 0.000 & 2.000 & 2.000 & 2.000 & 2.000 & 1.000 & 1.000 \\
        Min & -25.000 & 7.000 & 7.000 & 7.000 & 7.000 & 7.000 & 7.000 \\
        Kurtosis & -1.411 & -0.276 & -0.311 & -0.191 & -0.283 & -0.216 & -0.554\\
        Skewness & -1.477 & -0.547 & -0.160 & -0.196 & -0.081 & -0.410 & -0.395\\
    \bottomrule
    \end{tabular}
    }
    \caption{Human annotation statistics of DeepL.}
    \label{tab:stat-human-deepl}
\end{table}

\newpage
\phantom{.}
\newpage
\section{Average Correlation Coefficients between Human Evaluation and Automated Metrics}\label{sec:appendix-D}
See Table~\ref{tab:correlation}.
\begin{table}[h]
    \centering
    \begin{tabular}{lrrrr}
    \toprule
        & BLEU & chrF & COMET & BERTScore \\
    \midrule
        Error penalty & 0.141 & 0.147$^*$ & 0.170 & 0.185$^*$\\
        Coherence & 0.103 & 0.103 & 0.088 & 0.171\\
        Adherence to norms & 0.109 & 0.137 & 0.106 & 0.200$^*$\\
        Style, tone, and register appropriateness & 0.121 & 0.179 & 0.148 & 0.192\\
        Cultural sensitivity & 0.055 & 0.089 & 0.085 & 0.104\\
        Clarity & 0.085 & 0.134 & 0.095 & 0.150\\
        Practicality & 0.099 & 0.143 & 0.141 & 0.110\\
        Average & 0.102 & 0.133 & 0.119 & 0.159\\
    \bottomrule
    \end{tabular}
    \caption{Correlation coefficients between human evaluation and automated metrics. $^*$: $p<0.005$, significant in all six systems.}
    \label{tab:correlation}
\end{table}